\RequirePackage{etoolbox}
\csdef{input@path}%
{%
 {sty/},
 {img/},
}
\documentclass{elsevierbook}

\usepackage{natbib}
\usepackage{url}
\usepackage{tcolorbox}
\usepackage{multirow}

\usepackage{graphicx}
\graphicspath{ {img/} }

\definecolor{mygreen}{RGB}{115, 181, 179} 

\usepackage{tikz}
\newcommand*\circled[1]{\tikz[baseline=(char.base)]{
            \node[shape=circle,draw=none,inner sep=1pt, fill=mygreen,font=\sffamily] (char) {\textcolor{white}{\textbf{#1}}};}}

\usepackage[nameinlink]{cleveref}

\begin{document}

\Frontmatter

\begin{frontmatter}
\chapter{Open Challenges on Fairness of Artificial Intelligence in Medical Imaging Applications}

\begin{aug}
\author[]%
   {\fnm{Enzo} \snm{Ferrante}}%
\author[]%
   {\fnm{Rodrigo} \snm{Echeveste}}%

%
\end{aug}

\noindent \textit{\\Research Institute for Signals, Systems and Computational Intelligence, sinc(i)\\ CONICET, Universidad Nacional del Litoral\\
Argentina\\}
\minitoc

\begin{abstract}

Recently, the research community of computerized medical imaging has started to discuss and address potential fairness issues that may emerge when developing and deploying AI systems for medical image analysis. This chapter covers some of the pressing challenges encountered when doing research in this area, and it is intended to raise questions and provide food for thought for those aiming to enter this research field. The chapter first discusses various sources of bias, including data collection, model training, and clinical deployment, and their impact on the fairness of machine learning algorithms in medical image computing. We then turn to discussing open challenges that we believe require attention from researchers and practitioners, as well as potential pitfalls of naive application of common methods in the field. We cover a variety of topics including the impact of biased metrics when auditing for fairness, the leveling down effect, task difficulty variations among subgroups, discovering biases in unseen populations, and explaining biases beyond standard demographic attributes. 
\end{abstract}

  


\end{frontmatter}

\section{Introduction}\label{sec:intro}

The field of fairness of artificial intelligence (AI) for medical image analysis has rapidly grown during the last years. As it was discussed in the previous chapter, ensuring fairness in all the stages of the development and deployment of these technologies is crucial. Various sources of bias, including those related to data collection, model training, and clinical deployment, can impact the fairness of machine learning (ML) algorithms in medical image computing (MIC) \cite{ricci2022addressing}. Recent research has shown that AI systems can exhibit systematic bias against certain populations not only in tasks like face recognition \cite{buolamwini2018gender} or loan allocation, but also when dealing with healthcare data analysis in general, and medical images in particular \cite{seyyed2021underdiagnosis,larrazabal2020gender}. In this case, unique characteristics necessitate the adaptation of fairness notions to ensure equitable outcomes across different sub-populations. In this chapter, we aim at exploring and discussing some of the open challenges that we believe need to be addressed by the research community in the coming years. 

We start providing a brief recap on the reasons behind biased systems, highlighting potential challenges associated to every one of these aspects. We then continue discussing important aspects that we see as open challenges that we have found on fairness related research in medical image. We discuss a wide range of topics, including common pitfalls we have observed when auditing fairness with biased metrics (\Cref{sec::biased_metrics}); how can we address the leveling down effect that appears as a common consequence of bias mitigation strategies (\Cref{sec::leveling_down}); how differences in task difficulty for different subgroups may make it more challenging to audit for biases (\Cref{sec::different_difficulties}); the importance of discovering biases in unseen populations (\Cref{sec::unseen}) and how such biases may be explained beyond considering standard demographic attributes (\Cref{sec::beyond_standard}).

We stress the fact that this chapter is not intended to provide answers, but to ask questions and highlight problems that we have faced when doing research on fairness of AI for medical imaging during the last years, which we believe constitute important avenues for future research. 

\section{Why is my system biased?}\label{sec::why}

\begin{figure}[t!]
\centering
\includegraphics[width=0.7\linewidth]{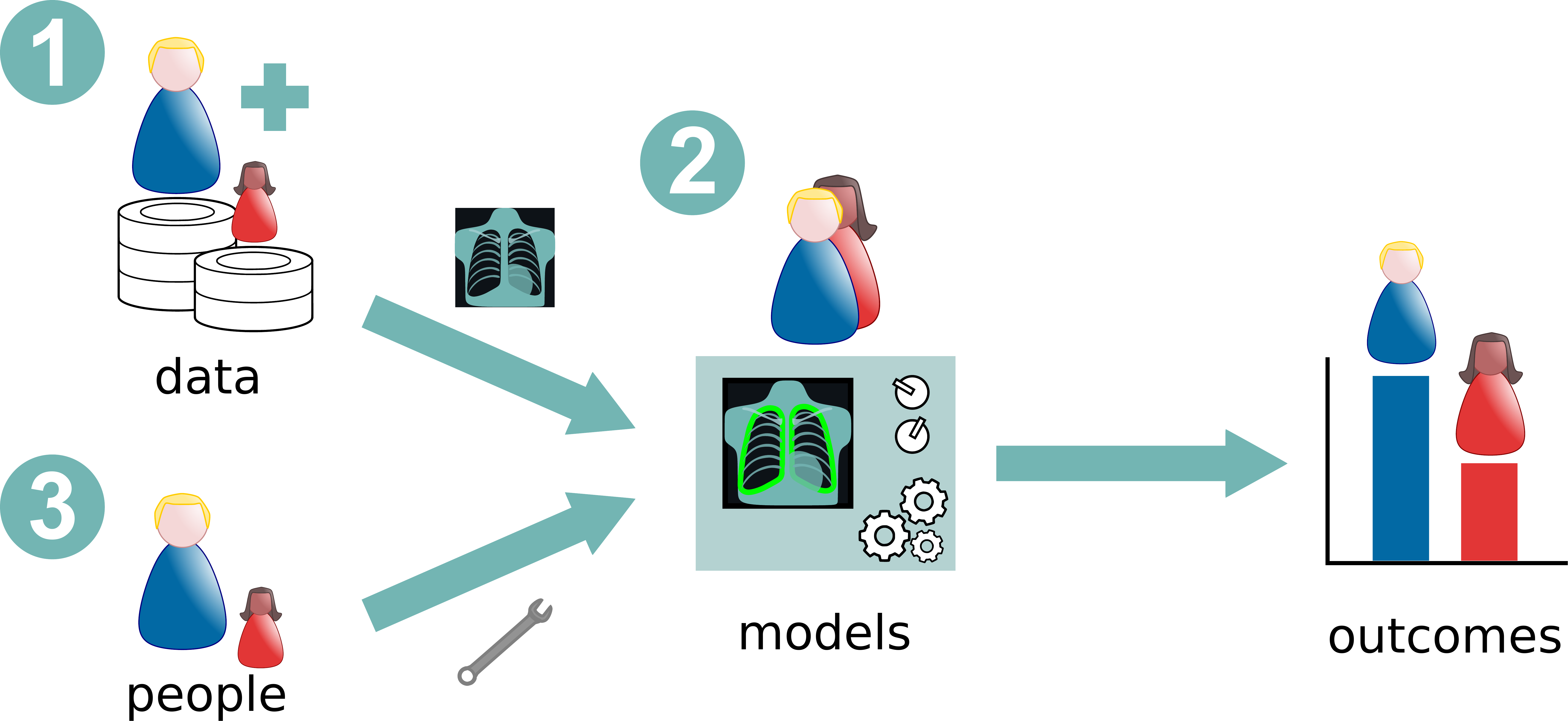}
\caption{
\textbf{Main potential sources of bias in AI systems for MIC.}
The data being fed to the system during training (1), design choices for the model (2), and the people who develop those systems (3), may all contribute to biases in AI systems for MIC. From Ref.~\cite{ricci2022addressing} with permission }\label{fig_sketch}
\end{figure}

As discussed in the previous chapter, biases in automated systems are multi-causal and may emerge at various stages of development; from the construction of the dataset to model deployment. When faced with such complexity, it often helps to partition the problem into broad categories. In this sense, sources of bias can be coarsely divided into: data, models and people (\Cref{fig_sketch}) \cite{ricci2022addressing} . While simplified, this grouping helps us build a mental picture of where biases might be coming from and how to intervene. So, when addressing fairness in ML models we should ideally be able to tick these three boxes. As we will see, important open challenges remain for each of these categories.  

\subsection{Data}
The first item on our list is \emph{data} (\Cref{fig_sketch}~\circled{1}). Data imbalance is a major source of bias for ML systems in MIC, often resulting in a lower performance for underrepresented groups \cite{larrazabal2020gender}. As commented in the previous chapter, biases might also emerge in this context due to differences in the properties of the imaging data between protected groups (domain shift), and to spurious correlations (via shortcut learning). Therefore, before we begin training a model, we should first ask ourselves if our training dataset is diverse enough to capture the richness of the target population at deployment. If multiple pathologies are present in the dataset, are there enough examples of each pathology for each sub-group? Intersectionality in terms of protected attributes should also be taken into account: how does the fraction of black females with condition A in the dataset compare to that of white males with the same condition? Differences in these fractions could lead to spurious correlations and shortcut learning \cite{banerjee2023shortcuts}, degrading model performance for under-represented sub-groups. Moreover, disparity in health access \cite{richardson2010access}, often means that social groups defined in terms of race, ethnicity, and income, tend to find medical help later, resulting in differences in the progression and severity of their condition at imaging time. So whenever possible, one must go beyond binary disease labels (find, no-find) and incorporate more detailed information into the analysis. Particular attention must be paid when an equivalent or higher performance is observed for under-represented groups. More severe stages of a disease may be simpler to detect, artificially inflating the performance of our system for underrepresented groups, hence masking biases (we will come back to the issue of varying difficulties in \Cref{sec::different_difficulties}). 

Data quality should also be assessed, especially when dealing with multi-site datasets. Differences in the equipment and imaging protocols may result in images of different quality. These factors again tend to correlate with income, affecting some groups more than others, and may be a source for spurious correlations (see Section \ref{sec::beyond_standard} for a more in-depth discussion about this topic). Finally, data balance does not end at the configuration of the dataset, and needs to be cared for throughout the development of the model: using data stratification to keep folds (and if possible even batches) balanced. Differences in data composition between train and validation sets, for instance, may bias the model if early-stopping is used to prevent over-fitting. Similarly, lack of batch stratification, may result in strong fluctuations in the representation of protected groups within the batch, resulting in less stable learning for these groups. We immediately see how the problem becomes exponentially complex and data-points quickly scarce for multiple protected groups, diseases, and acquisition sites, representing an open challenge in the field. 

\subsection{Models}
We next turn our attention to \emph{models} (\Cref{fig_sketch}~\circled{2}). Over the past decades, a variety of ML models have emerged for automated image processing, with deep learning models (DL) taking center stage in recent years, producing a veritable revolution across a wide range of fields. When we start working on a new problem, the choice of model class, complexity, regularization and training strategy will result in different inductive biases. These constitute implicit assumptions that will determine how a model will generalize from seen to unseen data, which we have seen lies at the core of biased systems via domain shift and shortcut learning, among others. While highly powerful tools, DL models present challenges in this regard. As mentioned in the previous chapter, DL models are more prone to shortcut learning, and less transparent than their rule-based AI predecessors (particularly in terms of their inductive biases). It is hence hard to predict when and how DL methods will fail when confronted with novel data. Furthermore, DL models tend to suffer from calibration problems, producing inaccurate uncertainty reports \cite{guo2017calibration}. Indeed, for out-of-distribution examples DL models may report a high degree of certainty while systematically failing. This may particularly affect unrepresented groups, and could be problematic in the medical domain, where multiple sources of information are usually combined to diagnose a pathology since optimal information fusion relies on the relative uncertainties of the individual sources \cite{bejjanki2011cue}. In this context, the challenge is how to develop schemes which retain the predictive power of DL models but incorporate inductive biases that favor generalization for under- or even unrepresented sub-populations. In chapter XX, the authors have presented a battery of solutions aiming at interventions before, during and after training of a model, to foster this type of generalization. Still much work remains to be done to explore which architectures provide benefits in each medical domain and image modality. In particular, transformer architectures which can incorporate multimodal information such as vision language models (VLMs) \cite{du2022survey}, constitute a promising avenue of research in the near future. Additionally, the use of multiple models constituting an ensemble, where predictions of individual models are collectively combined, has recently been proposed as a way to reduce biases in automated systems\cite{ko2023fair}.

\subsection{People}
Finally, we come to \emph{people} (\Cref{fig_sketch}~\circled{3}), by which be refer to those involved in the collection of data, model development and final deployment. It is now generally accepted that the AI community faces a structural diversity crisis including strong under representation in terms of gender and race \cite{freire2021measuring}. This lack of representation limits the ability of research groups to envision scenarios under which models should be audited for biases. Even with the best of intentions, it is simply harder for us to consider situations which are completely foreign to our reality. What's more, this lack of diversity is prevalent within the fairness in healthcare community itself, with an over-representation of white male authors from high-income countries, particularly when it comes to first and last authors \cite{alberto2024scientometric}. With the growth of international databases and benchmarks, open access libraries and foundational models, and the standardization of architectures and methods in ML, AI systems have an increasingly global impact. But what constitutes a protected group is often very specific to the local reality of a region, and is moreover fluid in time. It then becomes vital to involve the global community in data acquisition, model development and deployment. Racial bias in image detection has become a prototypical example of how even massively distributed software developed by major players in the AI ecosystem may present unforeseen problems at deployment. Indeed, almost a decade ago, a study showing how Google's photo app identified black individuals as gorillas made the headlines worldwide \footnote{https://archive.nytimes.com/bits.blogs.nytimes.com/2015/07/01/google-photos-mistakenly-labels-black-people-gorillas/}. It is hard to imagine a problem like this going unnoticed until deployment if a large number of developers working on and testing this system were black. In this sense, a report from the AI Now Institute reported in 2019 that only 2.5\% of Google's workforce was black, with other major companies such as Facebook and Microsoft reaching only 4\% \cite{west2019discriminating}. This is not to say that the problem is easy to solve, or that diversity by itself is a universal fix for biases. In fact, despite the notoriety of the case, and the resources available to these companies, eight years after Google prevented its software from categorizing any image as containing gorillas, an investigative report by the New York Times revealed that neither Google’s Photo App nor Apple’s can find gorillas, with similar problems arising for other primates and competing software from giants such as Amazon and Microsoft \footnote{https://www.nytimes.com/2023/05/22/technology/ai-photo-labels-google-apple.html}.

\section{Biased metrics may result in misleading fairness audits}\label{sec::biased_metrics}
An important challenge that arises during fairness audits, often overlooked, is the intrinsic bias that certain metrics may exhibit concerning properties varying across demographic groups. Imagine for example that we use the Dice coefficient \cite{dice1945measures} to measure the quality of a segmentation algorithm. Dice coefficient is known to penalize missed pixels more in small structures than in bigger structures \cite{reinke2024understanding}. In other words, it is biased with respect to the size of the structure of interest, as it tends to be lower in smaller objects. Consider auditing a segmentation model trained to segment organs in X-ray images, analyzing performance gaps across different age groups. Due to anatomical differences, some organs, such as the lung and heart, will be smaller in pediatric patients compared to adults. If the Dice coefficient is measured for both groups, and pediatric patients show lower scores, we must question if this discrepancy indicates a bias in the segmentation system or if it is a consequence of the Dice coefficient's inherent bias towards smaller structures. A similar concern may arise in fairness studies segmenting performance for male and female populations, where organ volumes are known to be different \cite{geraghty2004normal}. In these cases, adopting unbiased estimates of segmentation quality like the normalized Dice coefficient \cite{raina2023tackling} could potentially help to solve this issue. 

Another situation where such pitfalls may appear when auditing for fairness in medical image segmentation systems is related to strong differences in image resolution. Resolution is usually associated with factors such as the quality of the acquisition equipment and the duration of the acquisition protocol. Both of these factors depend on the resources locally available in a given health center, so that in low-resource settings images will most likely exhibit lower resolution. Let us say we have X-ray images coming from hospitals in countries with different income levels, which in turn present different resolutions. Our aim is to audit an anatomical segmentation system for fairness with respect to the geographic location, which in this case is a proxy for image resolution. After auditing, we observe a systematic lower Dice coefficient for images with lower resolution than for those with higher resolution. A lower resolution means that the same organ will be represented by less pixels than in the higher resolution image. Since Dice coefficient is biased by the 'size' of the structure in terms of pixels, most likely such bias will imply systematic lower values for the low resolution images than for the high resolution images. Thus, similarly as before, it will be difficult to determine if the Dice gap between both hospitals is a consequence of the metric bias or it really indicates lower segmentation quality in one population than the other. In this case, downsampling the high resolution images before computing the Dice coefficient may solve this particular problem. 

This is however not the end of the story, since biases could be further compounded if the prevalence of a condition varies from low to high resolution images, via shortcut learning. In general any spurious correlation with image resolution may result in biases during training. In this case downsampling before training may reduce these gaps, albeit potentially at the expense of a loss for the high resolution group. We go into further details about this in \Cref{sec::leveling_down} and \Cref{sec::beyond_standard}.

But image segmentation is not the only MIC task which may be affected by metric bias when auditing for fairness. In the context of image classification, an important aspect that is usually overseen is model calibration. Calibration, in essence, measures the reliability of prediction probabilities; a well-calibrated model produces probabilities that accurately reflect the true likelihood of outcomes. Recent work has shown that deep learning models, while highly effective in terms of discriminative performance, tend to present poorly calibrated posteriors \cite{guo2017calibration}. 

In the medical domain, quantifying predictive uncertainty is highly important for the interpretability of the outputs, in order to assist in clinical decision-making \cite{uncertaintyNature}. Moreover, if, as often is the case, multiple sources of information need to be combined in order to reach a diagnosis, optimal decision making relies on the relative uncertainties of the sources \cite{bejjanki2011cue}. In this context, miscalibrated models will result in higher error rates. Auditing for fairness in terms of calibration gaps for different demographic groups hence becomes highly important. Recently, however, it has been shown that standard metrics used to measure model calibration like the Expected Calibration Error (ECE) are heavily biased with respect to sample size, becoming monotonically worse with fewer samples in the evaluation set \cite{gruber2022better}. As discussed in recent work by Ricci Lara et al. \cite{ricci2023towards} such biased metrics could be misleading when analyzing ECE gaps in different sub-populations. Given that in algorithmic fairness studies, there is usually a majority group and an underrepresented minority group, if the metrics selected for the assessment are biased with respect to the sample size, it is possible that the results hide unfavorable behavior of AI models for the latter group. Again in this case the solution may be to look for unbiased estimators of calibration performance, or otherwise ensure that test sample sizes are the same across all sub-populations under analysis. Once again, this may not be trivial. If the smaller group has very few samples, such a reduction in sample sizes could lead to a substantial loss in statistical power.

\section{Fighting against the leveling down effect}\label{sec::leveling_down}
The current trend in machine learning fairness where researchers focus exclusively on improving a fairness metric or reducing the performance gap between different demographic groups may have negative consequences when resulting in the well known \textit{levelling down} effect. As discussed in recent work by Mittelstadt et al. \cite{mittelstadt2023unfairness}, the levelling down effect occurs when fairness is achieved at the expense of making every group worse off, or by bringing better performing groups down to the level of the worst performing ones. Such an outcome may arise after application of sophisticated methods, but could actually be readily achieved by a trivial solution. If we take for instance a binary classification model which performs better in group A than in group B, replacing a fraction of the predictions by a coin toss would reduce the gap. Indeed, as that fraction increases, both models would eventually reach by chance performance, being absolutely fair while also useless. One then needs to be careful when assessing the success of a bias mitigation scheme. Are we really doing better than a trivial solution? Levelling down effects have recently been shown to be a consequence of many popular methods to improve algorithmic fairness, both in computer vision \cite{zietlow2022leveling} and natural language processing tasks \cite{maheshwari2023fair}. This becomes especially important when dealing with medical data, as it has been postulated that such effect could be interpreted to violate the principles of bioethics, specifically that of non-maleficence \cite{ricci2022addressing}. 

In the context of healthcare, recent work discusses the impact of the leveling down effect and proposes a path towards achieving fairness via leveling up \cite{petersen2023path}. Imagine a computer assisted diagnosis system that shows higher performance for an adult population when comparing with a pediatric population. This can be seen as an unfair system, as it provides different standards of care to both groups. To reduce such gap in performance, the solution should then try to level up the worst group performance (i.e. in the pediatric group) instead of reducing the gap by delivering a worse standard of care to the adult population. However, achieving such a goal with standard bias mitigation methods is not straightforward, and it actually constitutes a very important challenge in the field. 

To overcome this limitation, the first step is to be aware of this issue and avoid reporting and analysing only performance gaps or worst group accuracy alone; we must also accompany this analysis by looking at average and subgroup model performance for both, advantaged and disadvantaged groups. There are many paths that can be considered to achieve fairness by leveling up performance in the disadvantaged group. As discussed in \cite{petersen2023path}, while one can work at the algorithmic and model level by implementing bias mitigation strategies, these will not always be effective when there are strong differences in task difficulty, as we will discuss in the next section.

A typical approach that is usually employed to improve model generalization (and may hence help mitigate biases against certain sub-populations) has to do with learning invariant representations. Since deep learning models are able to identify protected \cite{glocker2023algorithmic} characteristics even when such information is not provided (e.g. identifying race from X-ray images \cite{gichoya2022ai}), previous studies have warned about the potential risks associated to encoding protected characteristics in the learned representations. This information could potentially be leveraged for making predictions due to undesirable correlations in the (historical) training data. Thus, an active line of research for mitigating potential biases has to do with learning demographically invariant representations (e.g. via adversarial learning \cite{abbasi2020risk,li2021estimating, correa2021two}), which can then be used in downstream tasks like image classification or disease diagnosis. Even though this is a common trend, recent literature \cite{petersen2023demographically} challenges this assumption suggesting that approaches which aim at improving domain generalization by learning invariant representations (requiring models not to encode demographic attributes) may not be the best solution, as sometimes models may actually need to encode demographic-specific attributes. This may be the case for instance when there are strong differences in disease prevalence between the subgroups. 

In the same line, Weng and co-workers \cite{weng2024intentional} postulate that well intentioned attempts to prevent general purpose pretrained embeddings from learning sensitive attributes can have unintended consequences on the downstream models. They discuss how eliminating sensitive attributes from embeddings could potentially impair the performance of downstream models, particularly for historically marginalized patients, and suggest that assessing model bias should be conducted with the downstream model, rather than with the latent embeddings. Overall, the challenge of mitigating the leveling down effect persists as an unsolved issue, with no definitive solutions currently available.

\section{Is it biased or is it just more difficult?}\label{sec::different_difficulties}

Differences in task difficulty for different demographic subgroups have been previously reported when dealing with medical images. An example that is valid even for human radiologists is diagnosing thoracic diseases in X-ray images, as the shadow projected by the female breast may occlude a lung opacity or cause confusion during the interpretation of chest radiography, often leading to less effective examination of the lung's basal areas in women than men due to this factor \cite{alexander1958elimination}. Similar observations are made when analzying the results of X-ray classifiers trained with datasets which are imbalanced in terms of gender representation in \cite{larrazabal2020gender}. As discussed in \cite{petersen2023path}, even when the model was trained exclusively on data from women, its performance for certain diseases, like pleural thickening and pneumothorax, remained inferior in women compared to men, illustrating that a given task may be intrinsically harder in certain groups than others. Differences in breast density between sub-groups of women may also result in increased complexity when diagnosing breast cancer \cite{von2019sensitivity}. This may be an issue in fairness studies where the demographic attribute of interest, such as race or ethnicity, correlates with breast density. If a task is intrinsically more difficult for one demographic group than the other one due to anatomical differences, can we say that a model is unfair when it presents disparate performance for such groups? In \cite{ganz2021assessing}, the authors analyze this question both from a machine learning but also from an ethical perspective. Interestingly, they distinguish between cases where disparate performance originates in simple under-representation of a certain demographic group (i.e. data imbalance) from other cases where such disparities have more complex origins like differences in feature distribution or label noise among sub-population.

Although several studies that we have just mentioned point to anatomical differences as one of the potential causes behind biases and differences in task difficulty, recent work \cite{weng2023sex} challenges this statement with experimental results, which suggest that gender-based performance discrepancies in chest X-ray analysis, particularly in the widely utilized NIH and CheXpert Datasets, are due to dataset-specific factors rather than inherent physiological differences. Their analysis indicates that biological differences may not be the main driver of male–female performance gaps, pointing to other factors like distribution of various confounders, the prevalence of label errors, or differing recording quality as potential reasons.

\section{Discovering biases in unseen populations} \label{sec::unseen}
A recent study by Schrouff and colleagues \cite{schrouff2022diagnosing} investigated how fairness properties transfer across distribution shifts in the context of healthcare. Importantly, they found that, even if a system seems to be fair for a given source population with respect to a particular demographic attribute, when faced with a different target population which has been affected by some kind of distribution shift (e.g.  covariate shift, label shift, demographic shift, prevalence shift, etc) such fairness properties may not hold anymore. Imagine we audited a system designed to diagnose a certain disease in a pediatric hospital, and it turned out to exhibit similar performance for men and women, so we conclude it is not biased with respect to sex. In the next year, a new hospital which is not pediatric but a general one, adopts the same system. In this case, the new target population has suffered a demographic shift, as it now includes adult patients. Thus, we will need to perform a new audit to ensure that the system is still not biased, as shifts in intensity distribution, anatomical patterns and even prevalence shifts between pediatric and adult patients may have altered the fairness properties of our system.

One of the most common challenges in this scenario is the lack of ground-truth annotations in the new population where we want to use the system. As medical expert labels are expensive to obtain, when aiming to deploy a medical image analysis system in a new hospital, we most likely will not have access to such annotations. Thus, designing methods which can anticipate the existence of biases in unseen populations for which we do not have ground truth annotations becomes of paramount importance. This the case when deploying AI systems in new medical centers, especially if they were developed in a different country or region. A common example is the case of deploying AI systems in developing countries, which have been trained using international databases containing data coming mostly from developed countries. Demographic shifts between patients in different geographic locations, disparity of access between racial or ethnic groups, among other factors may end up decreasing predictive performance on a particular demographic group, and this may be overlooked.

Recent work suggests that, in certain tasks, it is possible to leverage automatic quality indices estimated in the absence of ground-truth to perform fairness audits and detect biases and gaps in performance. Let us take anatomical segmentation for example, a task prone to be biased due to under-representation of certain demographic groups \cite{puyol2021fairness}. In  \cite{gaggion2023unsupervised}, the authors propose a method based on the reverse classification accuracy framework (RCA) \cite{valindria2017reverse} to estimate performance gaps between demographic groups. RCA suggests building a reverse classifier that employs a single test image along with its predicted segmentation as a substitute for actual ground truth. Following this, the classifier is evaluated against a reference dataset that includes existing segmentations. The performance of this classifier is then utilized as an indirect measure of the predicted segmentation's quality. The experimental results included in \cite{gaggion2023unsupervised} show that such a simple method is enough to discover biases with respect to particular demographic attributes when performing anatomical segmentation. Methods such as these may be instrumental in anticipating biases, and devote limited research and auditing resources available in low-income countries to those factors which are most prone to produce biases.


\section{Explaining biases beyond standard demographic attributes}\label{sec::beyond_standard}

So far, we have mostly discussed auditing fairness and biases of medical AI systems with respect to standard demographic attributes like sex, gender, age or ethnicity. However, we believe an open challenge for the fairness community is related to re-thinking the way we audit biases, in particular how we define the demographic groups under analysis. An important work in this direction was the foundational paper by Joy Buolamwini and Timnit Gebru \cite{buolamwini2018gender}, which showed that auditing commercial face recognition systems from an intersectional demographic and phenotypical perspective, reveled strong biases against dark female individuals that would otherwise pass inadvertently if only aggregated results were reported. Such intersectional perspective implied defining groups of analysis by considering different protected attributes like gender and skin type at the same time. An important lesson learned from this foundational study is the role of group definitions: if one does not consider more complex attributes (in this case, defined as the combination of demographic and phenotypical attributes), then such biases would be overlooked.

In general, bias audits are performed following group fairness approaches, by selecting a pre-defined demographic attribute of interest, and comparing performance among sub-groups defined by such attribute. In the best case scenario, we also audit for intersectional groupings as discussed in \cite{buolamwini2018gender}, i.e. considering combinations of different demographic attributes. However, here we highlight the importance of auditing biases with respect to other types of attributes or potential confounders that go beyond standard demographic characteristics, and may be potential reasons for shortcut learning. In the context of medical images, other factors like anatomical properties (e.g. breast density, body mass index), hospitalization conditions (hospitalized vs ambulatory patients), hospitalization tubes, the presence of certain disease or even the brand and acquisition parameters of the imaging machine could potentially define failure modes for medical AI systems, since these features could be leveraged by machine learning models as shortcuts for predictions. If we audit employing groups of analysis based on standard demographic attributes only, we may end up overlooking potential biases most likely due to learned spurious correlations, ultimately delivering a poorer standard of care to such groups.

Similar problems have recently been addressed by the fairness literature in several ways. One of the first studies tackling the problem of fairness without demographics was presented by Lahoti and coworkers \cite{lahoti2020fairness}, who propose the Adversarially Reweighted Learning (ARL) strategy. ARL is implemented as a minimax game between a learner and adversary: while the learner optimizes for the main classification task, the adversary learns to identify regions where the learner makes significant errors, which can then be up-weighted in the loss function, forcing the learner to improve in these regions. This approach does not require any kind of demographic metadata, and can help to mitigate biases for unknown protected groups. A similar approach, which is based on uncertainty estimation instead of adversarial learning, was introduced by Stone and coworkers \cite{stone2022epistemic}. They propose using Bayesian neural networks with an epistemic uncertainty-weighted loss function to dynamically identify potential biases against individual training samples, and re-weighing them during training. In this study, the authors found a positive correlation between samples subject to bias and higher epistemic uncertainties. Similarly as before, this method allows to address biases without explicitly defining the groups of analysis, what could help to tackle the aforementioned problem, particularly when metadata (both for standard demographic attributes or other types of potential confounders) is not available. 

A different approach is adopted in \cite{kumar2023debiasing}, where the authors propose to leverage counterfactual image generation to explain the reasons behind biased systems, and combine this method with distributionally robust optimization (DRO) \cite{sagawa2019distributionally} to address biases caused by hospitalization artifacts. Using counterfactual explanations to understand biases is an interesting strategy, as it enables to provide visual feedback that could help in discovering failure modes beyond standard demographic attributes. Other strategies aiming at identifying biases using text-based explanations \cite{kim2023bias} generated by large vision language models \cite{du2022survey} have been recently proposed, but their validation in the context of medical imaging remains unexplored. These methods are not only useful during development but also after deployment since they increase transparency, which is an active demand from the medical community regarding ML models.

\section{Conclusion}

Throughout this chapter we have emphasized the importance of ensuring fairness in AI for medical imaging, and identified common pitfalls and open challenges that appear when auditing for biases. We have seen how biases can creep in at various stages of development and provided examples on how data, models and people may end up inadvertently contributing to biased systems. We also highlighted how the metrics we rely on to check if a system is fair might be biased themselves, making it tricky to truly assess how fair an AI system is. The challenges do not stop there: the variability in task difficulty among different demographic subgroups adds another layer of complexity, and there is no definitive answer as to how this issue should be addressed. Furthermore, the anticipation of biases in unseen and unlabeled populations underscores the importance of developing novel methodologies capable of producing early warnings, identifying and addressing biases preemptively. Moreover, the exploration of biases that transcend standard demographic attributes calls for a more sophisticated understanding of fairness and the way we define the demographic groups of analysis. Finally, while we have chosen to highlight these particular challenges based on our own experience and that of many of our colleagues, as well as the prominence we believe they have, it is important to acknowledge that our selection is inherently subjective. There are undoubtedly many other vital aspects of fairness in AI that merit attention and action. The journey towards trustworthy AI in medical imaging is complex and multifaceted, requiring us to continuously seek out and address new challenges as they arise.

\section*{References}

\Backmatter
%
%
%
%
%
%
\bibliographystyle{elsarticle-num}
%
%
%
\bibliography{biblio}

\begin{thebibliography}{10}
\expandafter\ifx\csname url\endcsname\relax
  \def\url#1{\texttt{#1}}\fi
\expandafter\ifx\csname urlprefix\endcsname\relax\def\urlprefix{URL }\fi
\expandafter\ifx\csname href\endcsname\relax
  \def\href#1#2{#2} \def\path#1{#1}\fi

\bibitem{ricci2022addressing}
M.~A. Ricci~Lara, R.~Echeveste, E.~Ferrante, Addressing fairness in artificial intelligence for medical imaging, nature communications 13~(1) (2022) 4581.

\bibitem{buolamwini2018gender}
J.~Buolamwini, T.~Gebru, Gender shades: Intersectional accuracy disparities in commercial gender classification, in: Conference on fairness, accountability and transparency, PMLR, 2018, pp. 77--91.

\bibitem{seyyed2021underdiagnosis}
L.~Seyyed-Kalantari, H.~Zhang, M.~B. McDermott, I.~Y. Chen, M.~Ghassemi, Underdiagnosis bias of artificial intelligence algorithms applied to chest radiographs in under-served patient populations, Nature medicine 27~(12) (2021) 2176--2182.

\bibitem{larrazabal2020gender}
A.~J. Larrazabal, N.~Nieto, V.~Peterson, D.~H. Milone, E.~Ferrante, Gender imbalance in medical imaging datasets produces biased classifiers for computer-aided diagnosis, Proceedings of the National Academy of Sciences 117~(23) (2020) 12592--12594.

\bibitem{banerjee2023shortcuts}
I.~Banerjee, K.~Bhattacharjee, J.~L. Burns, H.~Trivedi, S.~Purkayastha, L.~Seyyed-Kalantari, B.~N. Patel, R.~Shiradkar, J.~Gichoya, “shortcuts” causing bias in radiology artificial intelligence: causes, evaluation and mitigation., Journal of the American College of Radiology (2023).

\bibitem{richardson2010access}
L.~D. Richardson, M.~Norris, Access to health and health care: how race and ethnicity matter, Mount Sinai Journal of Medicine: A Journal of Translational and Personalized Medicine: A Journal of Translational and Personalized Medicine 77~(2) (2010) 166--177.

\bibitem{guo2017calibration}
C.~Guo, G.~Pleiss, Y.~Sun, K.~Q. Weinberger, On calibration of modern neural networks, in: International conference on machine learning, PMLR, 2017, pp. 1321--1330.

\bibitem{bejjanki2011cue}
V.~R. Bejjanki, M.~Clayards, D.~C. Knill, R.~N. Aslin, Cue integration in categorical tasks: Insights from audio-visual speech perception, PloS one 6~(5) (2011) e19812.

\bibitem{du2022survey}
Y.~Du, Z.~Liu, J.~Li, W.~X. Zhao, A survey of vision-language pre-trained models, arXiv preprint arXiv:2202.10936 (2022).

\bibitem{ko2023fair}
W.-Y. Ko, D.~D'souza, K.~Nguyen, R.~Balestriero, S.~Hooker, Fair-ensemble: When fairness naturally emerges from deep ensembling, arXiv preprint arXiv:2303.00586 (2023).

\bibitem{freire2021measuring}
A.~Freire, L.~Porcaro, E.~G{\'o}mez, Measuring diversity of artificial intelligence conferences, in: Artificial Intelligence Diversity, Belonging, Equity, and Inclusion, PMLR, 2021, pp. 39--50.

\bibitem{alberto2024scientometric}
I.~R.~I. Alberto, N.~R.~I. Alberto, Y.~Altinel, S.~Blacker, W.~W. Binotti, L.~A. Celi, T.~Chua, A.~Fiske, M.~Griffin, G.~Karaca, et~al., A scientometric analysis of fairness in health ai literature, PLOS Global Public Health 4~(1) (2024) e0002513.

\bibitem{west2019discriminating}
S.~M. West, M.~Whittaker, K.~Crawford, Discriminating systems, AI Now (2019) 1--33.

\bibitem{dice1945measures}
L.~R. Dice, Measures of the amount of ecologic association between species, Ecology 26~(3) (1945) 297--302.

\bibitem{reinke2024understanding}
A.~Reinke, M.~D. Tizabi, M.~Baumgartner, M.~Eisenmann, D.~Heckmann-N{\"o}tzel, A.~E. Kavur, T.~R{\"a}dsch, C.~H. Sudre, L.~Acion, M.~Antonelli, et~al., Understanding metric-related pitfalls in image analysis validation, Nature Methods (2024) 1--13.

\bibitem{geraghty2004normal}
E.~Geraghty, J.~Boone, J.~McGahan, K.~Jain, Normal organ volume assessment from abdominal ct, Abdominal imaging 29 (2004) 482--490.

\bibitem{raina2023tackling}
V.~Raina, N.~Molchanova, M.~Graziani, A.~Malinin, H.~Muller, M.~B. Cuadra, M.~Gales, Tackling bias in the dice similarity coefficient: Introducing ndsc for white matter lesion segmentation, arXiv preprint arXiv:2302.05432 (2023).

\bibitem{uncertaintyNature}
B.~Kompa, J.~Snoek, A.~L. Beam, Second opinion needed: communicating uncertainty in medical machine learning, NPJ Digital Medicine 4~(1) (2021) 1--6.

\bibitem{gruber2022better}
S.~Gruber, F.~Buettner, Better uncertainty calibration via proper scores for classification and beyond, Advances in Neural Information Processing Systems 35 (2022) 8618--8632.

\bibitem{ricci2023towards}
M.~A. Ricci~Lara, C.~Mosquera, E.~Ferrante, R.~Echeveste, Towards unraveling calibration biases in medical image analysis, in: MICCAI FAIMI Workshop 2023 (Fairness of AI in Medical Imaging), Springer, 2023, pp. 132--141.

\bibitem{mittelstadt2023unfairness}
B.~Mittelstadt, S.~Wachter, C.~Russell, The unfairness of fair machine learning: Levelling down and strict egalitarianism by default, arXiv preprint arXiv:2302.02404 (2023).

\bibitem{zietlow2022leveling}
D.~Zietlow, M.~Lohaus, G.~Balakrishnan, M.~Kleindessner, F.~Locatello, B.~Sch{\"o}lkopf, C.~Russell, Leveling down in computer vision: Pareto inefficiencies in fair deep classifiers, in: Proceedings of the IEEE/CVF Conference on Computer Vision and Pattern Recognition, 2022, pp. 10410--10421.

\bibitem{maheshwari2023fair}
G.~Maheshwari, A.~Bellet, P.~Denis, M.~Keller, Fair without leveling down: A new intersectional fairness definition, in: EMNLP 2023-The 2023 Conference on Empirical Methods in Natural Language Processing, 2023.

\bibitem{petersen2023path}
E.~Petersen, S.~Holm, M.~Ganz, A.~Feragen, The path toward equal performance in medical machine learning, Patterns 4~(7) (2023).

\bibitem{glocker2023algorithmic}
B.~Glocker, C.~Jones, M.~Bernhardt, S.~Winzeck, Algorithmic encoding of protected characteristics in chest x-ray disease detection models, Ebiomedicine 89 (2023).

\bibitem{gichoya2022ai}
J.~W. Gichoya, I.~Banerjee, A.~R. Bhimireddy, J.~L. Burns, L.~A. Celi, L.-C. Chen, R.~Correa, N.~Dullerud, M.~Ghassemi, S.-C. Huang, et~al., Ai recognition of patient race in medical imaging: a modelling study, The Lancet Digital Health 4~(6) (2022) e406--e414.

\bibitem{abbasi2020risk}
S.~Abbasi-Sureshjani, R.~Raumanns, B.~E. Michels, G.~Schouten, V.~Cheplygina, Risk of training diagnostic algorithms on data with demographic bias, in: Interpretable and Annotation-Efficient Learning for Medical Image Computing, Springer, 2020, pp. 183--192.

\bibitem{li2021estimating}
X.~Li, Z.~Cui, Y.~Wu, L.~Gu, T.~Harada, Estimating and improving fairness with adversarial learning, arXiv preprint arXiv:2103.04243 (2021).

\bibitem{correa2021two}
R.~Correa, J.~J. Jeong, B.~Patel, H.~Trivedi, J.~W. Gichoya, I.~Banerjee, Two-step adversarial debiasing with partial learning--medical image case-studies, arXiv preprint arXiv:2111.08711 (2021).

\bibitem{petersen2023demographically}
E.~Petersen, E.~Ferrante, M.~Ganz, A.~Feragen, Are demographically invariant models and representations in medical imaging fair?, arXiv preprint arXiv:2305.01397 (2023).

\bibitem{weng2024intentional}
W.-H. Weng, A.~Sellergen, A.~P. Kiraly, A.~D’Amour, J.~Park, R.~Pilgrim, S.~Pfohl, C.~Lau, V.~Natarajan, S.~Azizi, et~al., An intentional approach to managing bias in general purpose embedding models, The Lancet Digital Health 6~(2) (2024) e126--e130.

\bibitem{alexander1958elimination}
C.~Alexander, The elimination of confusing breast shadows in chest radiography, Australasian Radiology 2~(2) (1958) 107--108.

\bibitem{von2019sensitivity}
M.~von Euler-Chelpin, M.~Lillholm, I.~Vejborg, M.~Nielsen, E.~Lynge, Sensitivity of screening mammography by density and texture: a cohort study from a population-based screening program in denmark, Breast Cancer Research 21 (2019) 1--7.

\bibitem{ganz2021assessing}
M.~Ganz, S.~H. Holm, A.~Feragen, Assessing bias in medical ai, in: Workshop on Interpretable ML in Healthcare at International Connference on Machine Learning (ICML), 2021.

\bibitem{weng2023sex}
N.~Weng, S.~Bigdeli, E.~Petersen, A.~Feragen, Are sex-based physiological differences the cause of gender bias for chest x-ray diagnosis?, in: Workshop on Clinical Image-Based Procedures, Springer, 2023, pp. 142--152.

\bibitem{schrouff2022diagnosing}
J.~Schrouff, N.~Harris, S.~Koyejo, I.~M. Alabdulmohsin, E.~Schnider, K.~Opsahl-Ong, A.~Brown, S.~Roy, D.~Mincu, C.~Chen, et~al., Diagnosing failures of fairness transfer across distribution shift in real-world medical settings, Advances in Neural Information Processing Systems 35 (2022) 19304--19318.

\bibitem{puyol2021fairness}
E.~Puyol-Ant{\'o}n, B.~Ruijsink, S.~K. Piechnik, S.~Neubauer, S.~E. Petersen, R.~Razavi, A.~P. King, Fairness in cardiac mr image analysis: an investigation of bias due to data imbalance in deep learning based segmentation, in: Medical Image Computing and Computer Assisted Intervention--MICCAI 2021: 24th International Conference, Strasbourg, France, September 27--October 1, 2021, Proceedings, Part III 24, Springer, 2021, pp. 413--423.

\bibitem{gaggion2023unsupervised}
N.~Gaggion, R.~Echeveste, L.~Mansilla, D.~H. Milone, E.~Ferrante, Unsupervised bias discovery in medical image segmentation, in: MICCAI Workshop in Fairness of AI in Medical Imaging (FAIMI), Springer, 2023, pp. 266--275.

\bibitem{valindria2017reverse}
V.~V. Valindria, I.~Lavdas, W.~Bai, K.~Kamnitsas, E.~O. Aboagye, A.~G. Rockall, D.~Rueckert, B.~Glocker, Reverse classification accuracy: predicting segmentation performance in the absence of ground truth, IEEE transactions on medical imaging 36~(8) (2017) 1597--1606.

\bibitem{lahoti2020fairness}
P.~Lahoti, A.~Beutel, J.~Chen, K.~Lee, F.~Prost, N.~Thain, X.~Wang, E.~Chi, Fairness without demographics through adversarially reweighted learning, Advances in neural information processing systems 33 (2020) 728--740.

\bibitem{stone2022epistemic}
R.~S. Stone, N.~Ravikumar, A.~J. Bulpitt, D.~C. Hogg, Epistemic uncertainty-weighted loss for visual bias mitigation, in: Proceedings of the IEEE/CVF Conference on Computer Vision and Pattern Recognition, 2022, pp. 2898--2905.

\bibitem{kumar2023debiasing}
A.~Kumar, N.~Fathi, R.~Mehta, B.~Nichyporuk, J.-P.~R. Falet, S.~Tsaftaris, T.~Arbel, Debiasing counterfactuals in the presence of spurious correlations, in: Workshop on Clinical Image-Based Procedures, Springer, 2023, pp. 276--286.

\bibitem{sagawa2019distributionally}
S.~Sagawa, P.~W. Koh, T.~B. Hashimoto, P.~Liang, Distributionally robust neural networks for group shifts: On the importance of regularization for worst-case generalization, arXiv preprint arXiv:1911.08731 (2019).

\bibitem{kim2023bias}
Y.~Kim, S.~Mo, M.~Kim, K.~Lee, J.~Lee, J.~Shin, Bias-to-text: Debiasing unknown visual biases through language interpretation, ICML Workshop on Spurious Correlations, Invariance and Stability, 2023 2 (2023).

\end{thebibliography}
%
%
%
%
%
\end{document}